\documentclass[conference]{IEEEtran}
\IEEEoverridecommandlockouts
\usepackage{cite}

\usepackage{algorithm}
\usepackage{array}
\usepackage[caption=false,font=normalsize,labelfont=sf,textfont=sf]{subfig}
\usepackage{textcomp}
\usepackage{stfloats}
\usepackage{url}
\usepackage{verbatim}
\usepackage{graphicx}

\usepackage{balance}
\usepackage{caption}

\usepackage{graphicx}%
\usepackage{amsmath,amssymb,amsfonts}%
\usepackage{amsthm}%
\usepackage{mathrsfs}%
\usepackage{xcolor}%
\usepackage{textcomp}%
\usepackage{manyfoot}%
\usepackage{booktabs}%
\usepackage{algorithmicx}%
\usepackage{algpseudocode}%
\usepackage{listings}%
\usepackage{multirow}
\usepackage{color}
\usepackage{colortbl}
\usepackage{paralist}

\usepackage{hyperref}

\definecolor{mygray}{gray}{.95}
\definecolor{myblue}{RGB}{240,248,255}
\definecolor{myred}{RGB}{255,228,225}

\DeclareMathOperator{\EM}{E_{M}} 
\DeclareMathOperator{\EV}{E_{V}} 
\DeclareMathOperator{\ET}{E_{T}}

\DeclareMathOperator{\Sim}{Sim}

\def\ie{\emph{i.e.}}

\def\BibTeX{{\rm B\kern-.05em{\sc i\kern-.025em b}\kern-.08em
    T\kern-.1667em\lower.7ex\hbox{E}\kern-.125emX}}
\begin{document}

\title{T-MOR: Learning Motion-Aware Skeleton Representations for Human Action Recognition}

\author{%
  \IEEEauthorblockN{%
    Di Yang\textsuperscript{1} \quad
    Mahmoud Ali\textsuperscript{2} \quad
    Quan Kong\textsuperscript{3} \quad
    Gianpiero Francesca\textsuperscript{4} \quad
    François Brémond\textsuperscript{2}
    \vspace{0.cm}
  }
  \IEEEauthorblockA{%
  \textsuperscript{1}Suzhou Institute for Advanced Research, University of Science and Technology of China, Suzhou, China\\
    \textsuperscript{2}Inria Center at Université Côte d'Azur, Valbonne, France\\
    \textsuperscript{3}Woven by Toyota, Tokyo, Japan\quad
    \textsuperscript{4}Toyota Motor Europe, Brussels, Belgium
  }
}
\maketitle

\begin{abstract}
Vision-language models such as CLIP have recently shown promising results on a wide range of visual understanding tasks. However, most existing models rely primarily on appearance-level supervision from images or videos, and do not explicitly model human motion, which is essential for fine-grained and human-centric action recognition task as actions are defined by temporally structured and physically grounded body movements.
To address this problem, we propose Transferable skeleton MOtion Representation (T-MOR), a motion-aware framework that learns transferable action representations from skeleton sequences with the aid of video and language supervision during training. T-MOR adopts a multi-modal contrastive learning scheme that aligns skeleton motion with visual and textual representations, while performing inference using only lightweight skeleton inputs. To support large-scale pre-training, we construct PoseCap-1M, a new dataset that contains over one million synchronized video, skeleton, and text triplets covering diverse human activities.
We evaluate T-MOR on a range of human-centric action recognition benchmarks, including action classification and frame-wise temporal detection. Experimental results show that T-MOR consistently improves performance across multiple datasets, e.g., Toyota Smarthome, UAV-Human, Penn Action, TSU and Charades. In addition, T-MOR demonstrates strong generalization ability in few-shot and zero-shot settings, that highlights the effectiveness of motion-centric and embodied representations for transferable action understanding.
\end{abstract}


\section{Introduction}

Understanding human actions from videos is a crucial task in computer vision, with broad applications in healthcare monitoring, intelligent environments and robotics. Recent advances have been driven by three main directions: video-based models that learn spatio-temporal representations from RGB frames~\cite{Carreira_2017_CVPR, wang2023videomaev2, liu2025crcl, liu2025anomaly, liu2025privacy, li2026stnmamba} skeleton-based models that explicitly encode human motion through body joint trajectories~\cite{Yan2018SpatialTG, unik, motionbert2022, yang2024learning}, and vision-language foundation models (VLMs) that align visual content with natural language at scale~\cite{wang2023internvid, Ma2022XCLIP, ali2024, ali2024eccv}.

Despite their success, accurately understanding human actions in real-world videos remains challenging. Many actions are fine-grained, temporally dense, and multi-label, their recognition depends critically on precise modeling of human motion over time~\cite{Dai_2022_PAMI}. However, recent VLMs mainly rely on appearance and global semantics, and often lack explicit motion modeling, which limits their effectiveness on motion-centric recognition and temporal detection tasks. This gap reveals a key scientific challenge: how to leverage large-scale semantic supervision while preserving physically grounded and temporally coherent human motion representations.

\begin{figure}[t]
\begin{center}
\includegraphics[width=.98\linewidth]{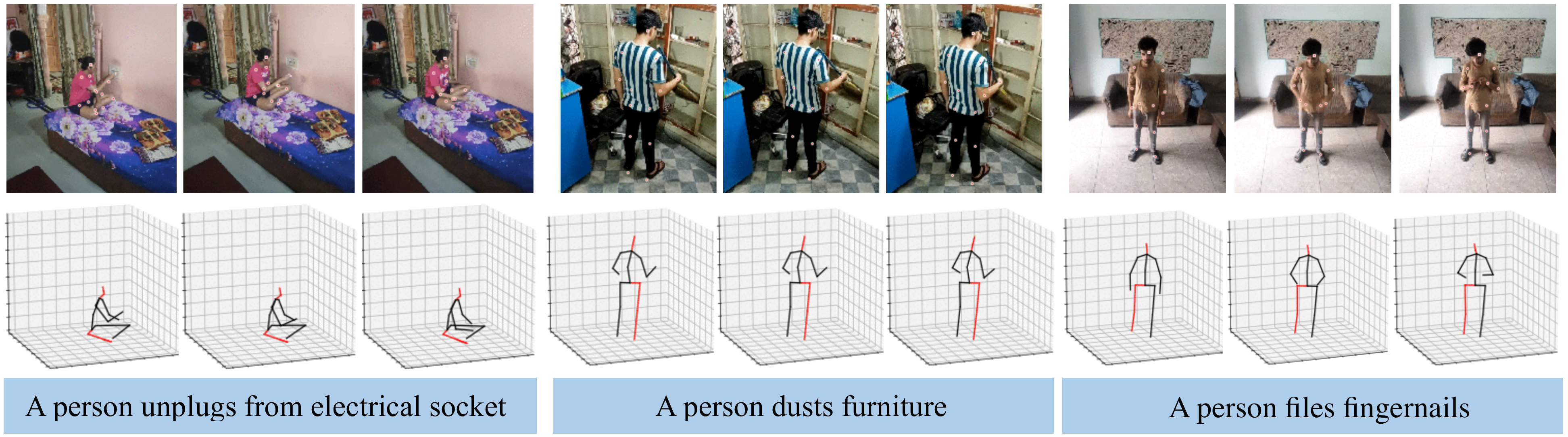}
\end{center}
   \vspace{-0.15cm}
   \caption{We propose a motion-aware human action representation learning framework with a pre-training dataset with multi-modal videos.}
\vspace{-0.2cm}
\label{fig:pose1m}
\end{figure}

From a human-centric embodied perception perspective, action understanding should be grounded in how humans move and interact with their environment, rather than relying solely on visual appearance. Skeleton representations provide a compact and explicit abstraction of embodied motion, which is robust to background clutter and visual variations (see Fig.~\ref{fig:pose1m}). However, most skeleton-based methods are trained in closed-set settings, this makes them difficult to transfer to new actions or tasks without extensive re-training.

\begin{figure*}[t]
\begin{center}
\includegraphics[width=.86\linewidth]{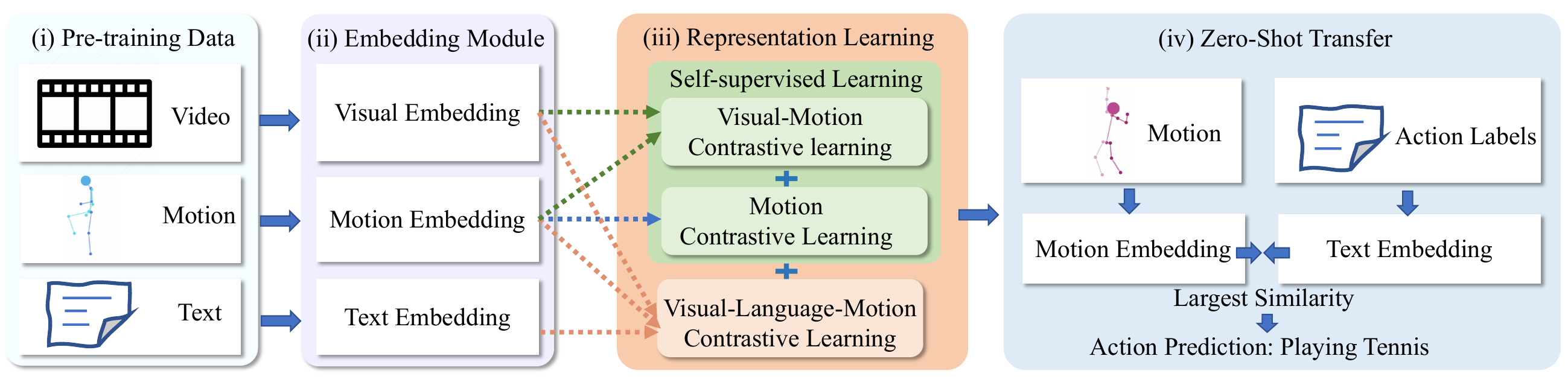}
\end{center}
   \vspace{-0.2cm}
   \caption{\textbf{T-MOR for zero-shot action recognition:} (i) The pre-training stage uses video, motion, and text inputs. (ii) The Embedding Module extracts multi-modal representations. (iii) The Motion Representation Learning stage combines Motion Contrastive Learning and Visual-Language-Motion Contrastive Learning to improve the motion model for complex human actions. (iv) In the Zero-Shot Transfer stage, the model predicts actions, such as "Playing Tennis," by finding the highest similarity between motion and text embeddings.}
\vspace{-0.15cm}
\label{fig:intro}
\end{figure*}

In this work, we propose a motion-aware framework that addresses this challenge for human action recognition. Our key idea is to use skeleton sequences as the core representation for action recognition, while exploiting visual and language modalities during training to enhance transferability. Based on this idea, we introduce a transferable human action representation framework, namely T-MOR, that aligns skeleton sequences (i.e., motion) with video and text embeddings through multi-modal contrastive learning. After pre-training, T-MOR performs inference using only lightweight skeleton inputs, so that it is suitable for real-world and resource-constrained scenarios (see Fig.~\ref{fig:intro}).
To further improve representation robustness, we incorporate a cluster-guided contrastive mechanism that exploits the intrinsic structure of the motion embedding space. The clustering process refines positive and negative pair selection by mitigating the issue of false negatives arising from random sampling in memory banks~\cite{He_2020_CVPR}, where many samples may belong to the same class as the query. By refining positive and negative sample selection during contrastive learning, this strategy encourages compact and discriminative motion representations without introducing additional model complexity.

To pre-train T-MOR, we construct a new multi-modal dataset, namely PoseCap-1M, to provide motion-video-text triplets. Then, we evaluate T-MOR on multiple benchmarks for real-world action classification and frame-level action detection. The results demonstrate that motion-aware and embodied representations can effectively complement visual-language supervision and lead to stronger generalization in challenging action understanding scenarios. We provide the related work section in Appendix.

Our contributions of this work can be summarized as follows:
\begin{inparaenum}[(i)]
\item We introduce T-MOR, a novel framework that learns transferable skeleton representations by aligning motion embedding with visual and language embeddings through cluster-guided multi-modal contrastive learning.
\item We collect PoseCap-1M, a large-scale action dataset to support the multi-modal pre-training of T-MOR.
\item Extensive Experiments show the effectiveness of T-MOR on both action classification and frame-wise temporal detection tasks, and highlight the importance of embodied motion modeling for human action understanding.
\end{inparaenum}

\section{Proposed Approach}

In this section, we introduce our proposed Transferable skeleton MOtion Representation learning architecture (T-MOR), the contrastive motion-video-language pre-training strategy and the PoseCap-1M pre-training dataset.

\subsection{Overview and Multi-modal Feature Extraction}

\begin{figure*}[t]
\begin{center}
\includegraphics[width=.86\linewidth]{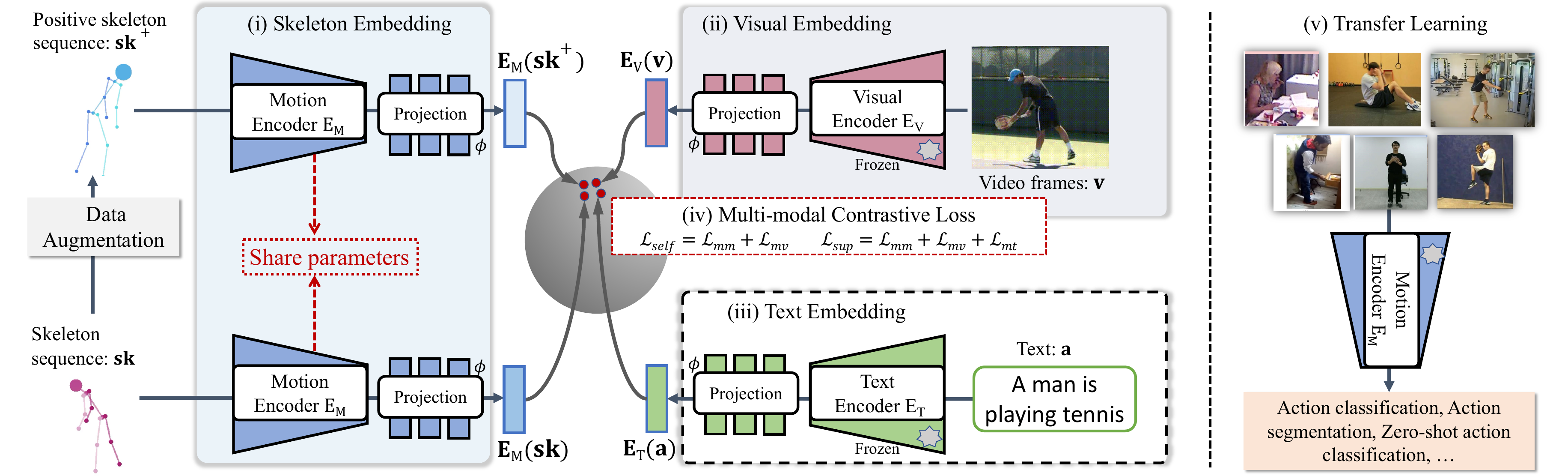}
\end{center}
   \vspace{-0.2cm}
  \caption{ \textbf{Overview of the T-MOR framework.} Given a skeleton sequence $\mathbf{sk}$, data augmentation is first applied to obtain $\mathbf{sk^+}$. The framework contains (i) Skeleton Embedding, where a motion encoder $\EM$ extracts human motion patterns; (ii) Visual Embedding, where a pre-trained encoder $\EV$ processes video frames $\mathbf{v}$ and provides visual cues; (iii) Text Embedding, where a pre-trained encoder $\ET$ encodes textual description $\mathbf{a}$ for action semantics. The three embeddings are passed through projection layers $\phi$ and then used in (iv) the Multi-modal Contrastive module, which jointly optimizes skeleton, visual, and text embeddings. Finally, (v) the pre-trained $\EM$ is transferred to downstream action recognition tasks.}
\vspace{-0.15cm}
\label{fig:overview}
\end{figure*}

As shown in Fig.~\ref{fig:overview}, we create a dataset (PoseCap-1M) that provides each clip as a triplet of skeleton motion, RGB video, and text description, denoted by $\mathbf{sk}$, $\mathbf{v}$, and $\mathbf{a}$. During pre-training, each skeleton sequence $\mathbf{sk}$ is augmented by random temporal cropping and random rotation to produce a positive motion view $\mathbf{sk^+}$. A motion encoder $\EM$ then maps the original and augmented sequences to $\EM(\mathbf{sk})$ and $\EM(\mathbf{sk^+})$. Unlike skeleton contrastive methods~\cite{ConNTU, guo2022aimclr, Mao_2023_ICCV} that rely only on augmented motion pairs, T-MOR contrasts motion features with both visual embeddings $\EV(\mathbf{v})$ and textual embeddings $\ET(\mathbf{a})$, which are produced by the frozen video encoder $\EV$ and text encoder $\ET$.
The goal is to learn a transferable skeleton encoder that remains usable with skeleton input alone at inference time. Thus, $\EM$ is optimized from scratch, and $\EV$ and $\ET$ are kept fixed throughout pre-training.
\vspace{0.15cm}

\noindent\textbf{Unified Skeleton Modeling and Embedding:}
We represent each skeleton sequence as a spatio-temporal tensor $\mathbf{sk}\in \mathbb{R}^{T \times J \times C_{in}}$, where $T$ denotes the number of frames, $J$ is the joint count, and $C_{in}$ is the coordinate dimension (2 for 2D skeletons and 3 for 3D skeletons). The motion branch is instantiated with UNIK~\cite{unik}, which captures both spatial joint relations and temporal evolution.

The motion encoder~\cite{unik} is a Transformer and TCNs combined architecture. 
After the processing of the motion encoder, a spatial-temporal pooling layer is performed to represent the skeleton sequence into a one-dimensional feature vector, denoted as $\EM(\mathbf{sk})$. For transferring $\EM$ to downstream tasks, we add a classifier on the spatio-temporal pooling layer for video-level classification or on a spatial pooling layer for frame-wise detection. 

\vspace{0.15cm}

\noindent\textbf{Video and Text Embedding:}
In this work, we apply ViCLIP~\cite{wang2023internvid} to extract video-language (video-textual) features $\EV(\mathbf{v})$ and $\ET(\mathbf{a})$, for given RGB video and the corresponding action label. 
ViCLIP~\cite{wang2023internvid} is a general video foundation model.
Its video branch is built on a spatio-temporal Vision Transformer (ViT)~\cite{vit}, and its language branch follows the Transformer text encoder design used in CLIP-style models~\cite{radford2021clip}. 
Rather than using ViCLIP features alone, we use them as semantic anchors for learning motion-aware skeleton representations. The video and text encoders remain frozen, while the projection heads and the skeleton motion encoder are trained with the proposed multi-modal contrastive objectives.

\subsection{Cluster-Guided Multi-modal Contrastive Learning}
T-MOR learns motion representations through two complementary training stages: 
a self-supervised motion-video pre-training stage and a supervised motion-text alignment stage. Both stages adopt our proposed cluster-guided contrastive mechanism, which refines the selection of positive and negative samples according to the semantic topology of the learned motion embedding space. Instead of uniformly sampling from the batch, we cluster the motion embeddings $\EM(\mathbf{sk})$ using DBSCAN~\cite{ester1996dbscan} after each epoch and form locally consistent groups. Within each cluster, samples are treated as positive pairs, while samples from other clusters serve as negatives. This structure-aware strategy helps reduce false negatives and encourages smooth manifold alignment across modalities.

\vspace{0.15cm}
\noindent\textbf{Self-supervised Motion-Video Pre-training:}
The first stage trains the motion encoder $\EM$ by exploiting motion-motion and
motion-video alignment signals without using any action annotations.
We first define the motion-motion contrastive objective as:
\begin{equation}\label{eq:self_contrast}
\small
\mathcal{L}_{mm}
= - \mathbb{E}\Big[
\log \frac{\mathcal{S}_{\text{align}}^{mm}}
{\mathcal{S}_{\text{align}}^{mm} + \mathcal{S}_{\text{disc}}^{mm}}
\Big],
\end{equation}
where $\mathcal{S}_{\text{align}}^{mm}$ measures the intra-cluster alignment
among semantically consistent motion samples, and
$\mathcal{S}_{\text{disc}}^{mm}$ enforces inter-cluster discrimination
against samples from other clusters:
\begin{equation}\label{contras-lossmm}
\small
    \mathcal{S}_{\text{align}}^{mm} =  
    \sum_{\mathbf{sk}^+ \in \mathcal{P}_{\mathbf{sk}}}
    e^{\Sim\big ( \EM(\mathbf{sk}), \EM (\mathbf{sk}^+) \big ) },
\end{equation}
\begin{equation}    
\small
    \mathcal{S}_{\text{disc}}^{mm} =  
    \sum_{\mathbf{sk}^- \in \mathcal{N}_{\mathbf{sk}}}
    e^{\Sim\big( \EM(\mathbf{sk}), \EM(\mathbf{sk}^-) \big ) }.
\end{equation}

Similarly, the motion-video contrastive objective $\mathcal{L}_{mv}$ is defined
in the same form by replacing $\EM(\mathbf{sk}^+)$ and $\EM(\mathbf{sk}^-)$ with
the visual representations $\EV(\mathbf{v})$ and
$\EV(\mathbf{v}^-)$.
Here, $\mathcal{P}_{\mathbf{sk}}$ and $\mathcal{N}_{\mathbf{sk}}$ denote the
intra- and inter-cluster sets determined by DBSCAN on top of all the motion embeddings in the dataset. $\mathbf{v}$ is the corresponding visual feature of the same sequence of $\mathbf{sk}$ and $\mathbf{v}^-)$ is the visual feature of each negative sample $\mathbf{sk}^-$. $\mathbf{sk}^-$ is randomly selected from the $\mathcal{N}_{\mathbf{sk}}$.
The similarity function $\Sim(\cdot)$ denotes the cosine similarity with
temperature scaling~\cite{non-para}, formulated as:
\begin{equation}\label{sim}
\small
    \Sim (x, y) =  
    \frac{\phi (x) \cdot \phi (y)}
    {\left\| \phi (x) \right\| \cdot \left\| \phi (y) \right\|} 
    \cdot \frac{1}{Temp},
\end{equation}
where $Temp$ is the temperature hyper-parameter, and $\phi$ represents a
learnable projection head (\ie, an MLP) that improves representation smoothness
and modality alignment.
The total self-supervised objective is defined as:
\begin{equation}
\small
    \mathcal{L}_{self} = \mathcal{L}_{mm} + \mathcal{L}_{mv}.
\end{equation}
This stage learns a transferable skeleton motion encoder that is explicitly
aligned with visual motion cues, providing a robust initialization for
subsequent training stages.

\vspace{0.15cm}
\noindent\textbf{Supervised Motion-Text Alignment:}
In the supervised setting, we perform a second training stage to align motion
representations with language semantics, enabling semantic understanding and
zero-shot transfer.
The same cluster-guided contrastive formulation is adopted to preserve the
structural consistency of motion-text alignment.
The supervised objective is defined as:
\begin{equation}\label{eq:sup_contrast}
\small
\mathcal{L}_{mt}
= - \mathbb{E}\Big[
\log \frac{\mathcal{S}_{\text{align}}^{mt}}
{\mathcal{S}_{\text{align}}^{mt} + \mathcal{S}_{\text{disc}}^{mt}}
\Big],
\end{equation}
where
\begin{equation}\label{contras-lossmt}
\small
    \mathcal{S}_{\text{align}}^{mt} =  
    e^{\Sim\big ( \EM(\mathbf{sk}), \ET(\mathbf{a}) \big ) },
\end{equation}
\begin{equation}    
\small
    \mathcal{S}_{\text{disc}}^{mt} =  
    \sum_{\mathbf{sk}^- \in \mathcal{N}_{\mathbf{a}}}
    e^{\Sim\big( \EM(\mathbf{sk}), \ET(\mathbf{a}^-) \big ) }.
\end{equation}
Similarly, $\mathbf{a}$ is the corresponding text feature of the same sequence of $\mathbf{sk}$ and $\mathbf{a}^-$ is the text feature of each negative sample $\mathbf{sk}^-$ which is randomly selected from the $\mathcal{N}_{\mathbf{sk}}$.

In summary, T-MOR introduces a unified cluster-guided contrastive framework for
both self-supervised and supervised training settings:
\begin{equation}\label{eq:total_loss}
\mathcal{L}_{sup} = \mathcal{L}_{mm} + \mathcal{L}_{mv} + \mathcal{L}_{mt}.
\end{equation}
By integrating the cluster-guided mechanism directly into each contrastive
objective, our approach reformulates the conventional InfoNCE paradigm into a
structure-aware multi-modal optimization. This design encourages motion
embeddings to form semantically compact yet discriminative manifolds across
skeleton, visual, and textual modalities, leading to stronger generalization
and improved interpretability without introducing additional model complexity.

\subsection{PoseCap-1M: Large-scale Training Dataset}
Transferable vision foundation models are typically learned from large image-text~\cite{radford2021clip} or video-text corpora~\cite{wang2022internvideo, xu2021vlm-videoclip}. An analogous resource is needed for skeleton motion: a large human-centric collection that jointly provides videos, language, and skeleton sequences. Existing public datasets do not yet offer this combination at sufficient scale.

Indoor laboratory datasets~\cite{Shahroudy2016NTURA, NTU-120} often lack the occlusions, compositional activities, and viewpoint changes that appear in unconstrained videos, which limits their value for pre-training broadly transferable models. To fill this gap, we construct PoseCap-1M as a large-scale skeleton-centered pre-training dataset (see Fig.~\ref{fig:pose1m}). The Appendix provides further details on data collection.

\begin{table*}[t]
\centering

\begin{center}
\scalebox{0.88}{
\setlength{\tabcolsep}{3mm}{
\begin{tabular}{  l c c c c c c c}

\hline
\multirow{2}*{\textbf{Methods}}
& \multirow{2}*{\textbf{Pre-train}}& \multirow{2}*{\textbf{Boackbone}}& \multicolumn{2}{c}{\textbf{ Smarthome}} & \multicolumn{2}{c}{\textbf{UAV-Human}}& \multicolumn{1}{c}{\textbf{Penn Action}}\\

& & & \text{CS(\%)} 
&\text{CV2(\%)}& {\text{ CS1(\%)}} & {\text{CS2(\%)}} & {\text{Top-1(\%)}} \\
\hline
\hline
\rowcolor{mygray}\text{Previous SoTA~\cite{yang2022via}}& Posetics& UNIK &51.9& 52.2& 32.9 &56.1 & 97.3\\
\text{UNIK(Backbone)~\cite{unik}}&None &UNIK&\text{24.6} & 20.7 &3.8  &4.1 &29.8 \\

\text{T-MOR (Ours)}& PoseCap-1M (V+M)& UNIK&\text{49.3}& \text{46.4} &\text{27.6} &\text{43.4}  &\text{86.3} \\

\textbf{T-MOR (Ours)}& PoseCap-1M (V+M+T)& UNIK &\textbf{52.6}& \textbf{53.4} & \textbf{33.5}&\textbf{60.1} & \textbf{97.8} \\

\hline
\rowcolor{mygray}\text{3s-CrosCLR~\cite{ConNTU}}&Posetics& ST-GCN &\text{52.6} &\text{58.1} & \text{40.0} & -&\text{95.2}\\
\rowcolor{mygray}\text{AimCLR~\cite{guo2022aimclr}}&Posetics &AGCN&\text{62.7} &\text{61.1} & \text{40.5} &- &\text{96.6}\\

\rowcolor{mygray}\text{MotionBert~\cite{motionbert2022}}&Human3.6M&Transformer&\text{64.2} &\text{63.7} & \text{-} &-  &\text{-}\\

\rowcolor{mygray}\text{ViA~\cite{yang2022via}}&Posetics &UNIK&\text{64.5} &\text{65.2} & \text{42.6} &69.5 &\text{98.0}\\
\rowcolor{mygray}\text{ViA~\cite{yang2022via}}&PoseCap-1M &UNIK&\text{64.9} &\text{65.8} & \text{43.3} & -&\text{-}\\

\text{T-MOR (Ours)}&PoseCap-1M (V+M)& UNIK&\text{63.2} &\text{61.8}  &\text{40.4} &\text{67.8} &\text{96.2} \\

\textbf{T-MOR (Ours)}& PoseCap-1M (V+M+T)  &UNIK &\textbf{66.2} &\textbf{66.7} & \textbf{44.4} &\textbf{70.8} &\textbf{98.2}\\

\hline
\end{tabular}}}

\end{center}
\vspace{-0.3cm}
\caption{\textbf{Action classification} results on Smarthome, UAV-Human, and Penn Action after pre-training, evaluated by \text{linear probing (top)} and \text{fine-tuning (bottom)}. M/V/T: Motion/Visual/Text. Gray rows report SoTA results using skeletons.}
\vspace{-0.cm}
\label{tab_tr_cls}
\end{table*}

\begin{table*}[t]
\centering

\begin{center}
\scalebox{0.88}{
\setlength{\tabcolsep}{4mm}{
\begin{tabular}{  l c c c c c c c c }

\hline
\multirow{2}*{\textbf{Methods}}
& \multirow{2}*{\textbf{Pre-train}}& \multirow{2}*{\textbf{Backbone}}& \multicolumn{3}{c}{\textbf{TSU}} & \multicolumn{2}{c}{\textbf{Charades}}\\
& & &\text{\#Params}& \text{CS(\%)} &\text{CV(\%)} &\text{\#Params}& {\text{mAP(\%)}} \\
\hline
\hline
\rowcolor{mygray}\text{Previous SoTA~\cite{Dai_2022_PAMI}}& PoseCap-1M (M)&UNIK &-&\text{15.2}& \text{18.1} & - &\text{7.7}\\
\text{UNIK (Backbone)~\cite{unik}}&None & UNIK&13.1K&\text{8.1}& \text{6.9}&  40.2K &6.1 \\

\text{T-MOR (Ours)}& PoseCap-1M (V+M)& UNIK& 13.1K&\text{19.8} & \text{12.6} & 40.2K&\text{11.3} \\

\textbf{T-MOR (Ours)}& PoseCap-1M (V+M+T)& UNIK& 13.1K&\textbf{23.2}&  \textbf{19.4}&   40.2K& \textbf{16.6} \\

\hline

\rowcolor{mygray}\text{Bi-LSTM~\cite{graves2005framewise}}&Smarthome& LSTM&- &\text{17.0} &14.8 &-&8.2 \\
\rowcolor{mygray} \text{TGM~\cite{TGM}}&Smarthome &TCN &- &\text{26.7} &13.4 &-&9.0 \\
\rowcolor{mygray}\text{SD-TCN~\cite{Dai_2022_PAMI}}&Smarthome& AGCN&- &\text{26.2}&22.4 &-&9.8 \\
\rowcolor{mygray}\text{SD-TCN~\cite{Dai_2022_PAMI}}&PoseCap-1M (M)&UNIK &- &\text{31.2}&22.8 &-&16.5 \\
UNIK (Backbone)~\cite{unik} & None&UNIK &3.45M &28.2 &11.0 &3.45M &18.6  \\
\text{T-MOR (Ours)}& PoseCap-1M (V+M)&UNIK  &3.45M&\text{33.4}& \text{{21.9}} &3.45M &\text{18.3} \\

\textbf{T-MOR (Ours)}&PoseCap-1M (V+M+T) &UNIK &3.45M&\textbf{38.3}& \textbf{23.6} & 3.45M &\textbf{26.0}\\

\hline
\end{tabular}}}

\end{center}
\vspace{-0.3cm}
\caption{\textbf{Action detection} results on Toyota Smarthome Untrimmed (TSU) and Charades after pre-training, evaluated by \text{linear probing (top)} and \text{fine-tuning (bottom)}. M/V/T: Motion/Visual/Text. Gray rows report SoTA skeleton-based results.}
\vspace{-0.15cm}
\label{tab_tr_seg}
\end{table*}  

\section{Experiments and Analysis}

We evaluate T-MOR on multiple datasets with extensive experimental analyses. 
Firstly, we study the generalization ability of T-MOR by quantifying the performance improvement obtained by transfer-learning on real-world 2D action recognition after visual-motion and visual-motion-text pre-training on the \textbf{PoseCap-1M} dataset. Specifically, we use \textbf{Toyota Smarthome, UAV-Human} and \textbf{Penn Action} for action classification and \textbf{TSU} and \textbf{Charades} for action detection.
Secondly, we evaluate zero-shot transfer and 3D action detection after visual-motion-text pre-training. Notably, we only use skeleton data and the skeleton encoder on downstream tasks.
Finally, we analyze the contribution of the main design choices through ablation experiments. 
See Appendix 
for dataset description, implementation details, and additional studies.

\subsection{Evaluation on Action Classification}\label{sec:classification}
We evaluate T-MOR on action classification with two transfer settings: \textit{linear probing}, in which the backbone is fixed and only the fully-connected classifier is learned, and \textit{fine-tuning}, in which all network parameters are updated. After pre-training on PoseCap-1M, we transfer the motion encoder $\EM$ to three 2D skeleton-based classification benchmarks, namely \text{Smarthome, UAV-Human} and \text{Penn Action}, without using video or text inputs during evaluation.
\vspace{0.1cm}

\noindent{\textbf{Linear Probing and Fine-tuning:}}
Tab.~\ref{tab_tr_cls} (top) presents the linear probing results on the three 2D datasets. Since only the classifier is trained in this setting, the comparison with a randomly initialized backbone directly reflects the quality of the pre-trained representation. Both motion-visual and motion-visual-text pre-training lead to clear improvements, especially on smaller datasets, such as +32.7\% on Smarthome CV2 and +68.0\% on Penn Action compared with training from scratch. This shows that the learned skeleton encoder can provide transferable action features from skeleton sequences.
Tab.~\ref{tab_tr_cls} (bottom) presents the fine-tuning results, where the full network is optimized on each target dataset. The results show that our pre-training can further improve over previous skeleton-based SoTA methods~\cite{ConNTU, guo2022aimclr, motionbert2022, yang2022via} that also use a pre-training stage, for example +1.5\% on Smarthome CV2. To make the comparison fair, we also re-implement the SoTA method~\cite{yang2022via} with the same backbone and the same pre-training data. The self-supervised motion-visual model remains competitive with supervised pre-trained models~\cite{yang2022via, motionbert2022}. These results suggest that large-scale video data can benefit downstream skeleton-based action classification even when action labels are not used during pre-training.

\subsection{Evaluation on Action Detection}\label{sec:detection}

We further evaluate T-MOR on action detection with the same two transfer settings, \textit{linear probing} and \textit{fine-tuning}. The experiments are conducted on two action detection benchmarks, \text{TSU} and \text{Charades}, after pre-training on \text{PoseCap-1M}.

\vspace{0.1cm}
\noindent{\textbf{Linear Probing and Fine-tuning:}}
Tab.~\ref{tab_tr_seg} (top) reports the linear probing results on the two 2D detection datasets. Compared with training the same backbone from scratch, full motion-visual-text pre-training brings substantial gains, such as +15.1\% on TSU CS and +10.5\% on Charades. This demonstrates that the pre-trained motion encoder can produce useful skeleton features for long and complex temporal detection tasks. The motion-visual pre-training setting also improves transfer performance, which shows that RGB supervision is useful when action annotations are not available.
Tab.~\ref{tab_tr_seg} (bottom) reports the fine-tuning results. The motion-visual-text pre-trained model outperforms supervised skeleton-based SoTA methods~\cite{graves2005framewise, TGM, Dai_2022_PAMI}, with gains such as +11.6\% on TSU CS and +16.2\% on Charades. For a fair comparison, we re-implement the SoTA method~\cite{Dai_2022_PAMI} using the same backbone and the same pre-training dataset. The self-supervised motion-visual pre-training setting also achieves competitive results.

\subsection{More Analysis on Transfer-learning}

\begin{table}[t]
\centering

\begin{center}
\scalebox{.9}{

\setlength{\tabcolsep}{2.8mm}{
\begin{tabular}{  l c c c }

\hline
\multirow{2}*{\textbf{Methods}} & 
\multicolumn{2}{c}{\textbf{Smarthome}} & {\textbf{Penn}}\\
& \text{CS(\%)} &\text{CV2(\%)} & {\text{Top-1(\%)}} \\
\hline
\hline

\text{CLIP~\cite{radford2021clip}}& \text{10.1}& \text{13.6}  & \text{63.1}\\
\text{XCLIP~\cite{Ma2022XCLIP}} &\text{16.5}& \text{14.8} & 72.7\\
\text{ViCLIP~\cite{wang2023internvid}} &\text{15.4}& \text{14.6} & 74.3\\
\text{MotionCLIP~\cite{tevet2022motionclip}}&  2.6& \text{2.2} & 6.1\\

\hline 
\text{Video-LLaVA~\cite{lin2023videollavalearningunitedvisual}} &\text{7.2}& \text{2.5}  & \text{60.5} \\
\text{LLaVA-OneVision~\cite{li2024llavaonevisioneasyvisualtask}} &\text{8.0}& \text{6.4}  & \text{57.1} \\
\text{LAVIDAL~\cite{chakraborty2024llavidal}} &\text{9.1}& \text{6.8}  & \text{43.4} \\
\text{Video-Chatgpt~\cite{Maaz2023VideoChatGPT}} &\text{6.0}& \text{2.4}  & \text{13.4} \\

\hline 
\text{T-MOR (Motion only)} &\text{14.5}& \text{7.0}  & \text{69.5} \\
\textbf{T-MOR (Motion+Visual)} &\textbf{21.9}& \textbf{17.4}  & \textbf{80.9} \\
 
\hline
\end{tabular}}}

\end{center} 
\vspace{-0.3cm}
\caption{\textbf{Zero-shot} transfer results and comparisons on Smarthome (Top-1 accuracy) and Penn Action action classification benchmarks without re-training.}
\vspace{-0.15cm}
\label{tab_zero}
\end{table}  

\noindent\textbf{Zero-shot Transfer:}\label{sec:zero-shot}
The zero-shot transfer capabilities of video-language model are very important to showcase their generalizability to unseen actions, leveraging the knowledge gained from the pre-training without direct training on specific action labels. In this study, we provide the comparisons in Tab.~\ref{tab_zero} between recent SoTA CLIP-based models~\cite{Ma2022XCLIP, wang2023internvid} (top) and Video Question Answering (VQA) models~\cite{lin2023videollavalearningunitedvisual, li2024llavaonevisioneasyvisualtask, chakraborty2024llavidal, Maaz2023VideoChatGPT} (middle) that can provide zero-shot results on challenging real-world action classification datasets, Smarthome and Penn Action. For CLIP-based models including T-MOR, we follow~\cite{radford2021clip} to utilize textual descriptions of actions as proxies for action classes, to enable these models to predict actions in videos on which they have not been trained. 
For VQA methods, we directly generate predictions of the action for a video clip by asking questions about the action in the video (the action list of the dataset is given in the question). The results demonstrate that with video-language model, CLIP-based model~\cite{wang2023internvid, Ma2022XCLIP} are more general for zero-shot action classification.

Subsequently, we test whether T-MOR can perform zero-shot recognition from skeletons alone on real-world action classification benchmarks. Under this setting, T-MOR outperforms the existing skeleton model~\cite{tevet2022motionclip} trained with visual-textual features. From the results in Tab.~\ref{tab_zero} (bottom), T-MOR 
performs competitively with the current video-language foundation models~\cite{wang2023internvid, Ma2022XCLIP}. This shows its potential for practical applications when training data are not available. Moreover, we show that motion features are complementary to visual-language features, by combining both motion and RGB features from T-MOR by average pooling. This can be the most effective way to achieve SoTA accuracy for zero-shot applications.

\vspace{0.1cm}

\noindent\textbf{Transfer Ability on 3D Action Detection:}\label{sec:3d-detection}
We further test T-MOR on PKU-MMD to verify whether the learned skeleton motion representation can transfer to 3D action detection. As shown in Tab.~\ref{tab_pku}, T-MOR achieves strong event-level mAP at different IoU thresholds. These results show that the proposed pre-training is not limited to 2D real-world benchmarks and can also benefit 3D skeleton detection.
\begin{table}[t]
\centering

\begin{center}
\scalebox{0.95}{
\setlength{\tabcolsep}{2.0mm}{
\begin{tabular}{ l c c c c }
\hline
\multirow{2}*{\textbf{Methods}}& \multirow{2}*{\textbf{Modality}}& \multicolumn{3}{c}{\textbf{PKU-MMD} mAP@IoU} \\
&&{\text{ 0.1(\%)}} & {\text{0.3(\%)}} & {\text{0.5(\%)}}\\

\hline
\hline
\rowcolor{mygray}GRU-GD~\cite{GRU-GD} & RGB &82.4& 81.3& 74.3\\
\rowcolor{mygray}SSTCN-GD~\cite{Dai_2021_ICCV} & RGB&83.7 &82.1 &76.5\\
\rowcolor{mygray}Augmented-RGB~\cite{Dai_2021_ICCV} & RGB&86.3 &84.5 &81.1\\

\hline
JCRRNN~\cite{jcrnn} &Skeleton& 45.2& \text{-} &32.5 \\
Convolution Skeleton~\cite{liu2017pku} & Skeleton&49.3& 31.8 &12.1 \\
Skeleton boxes~\cite{skeletonbox}& Skeleton& 61.3& \text{-} & 54.8\\
Window proposal~\cite{windowproposal} &Skeleton &92.2& \text{-} &90.4 \\

\hline
\textbf{T-MOR (Ours)}&Skeleton &\textbf{94.3} &\textbf{93.2} & \textbf{90.7} \\
\hline
\end{tabular}}}
\end{center}
\vspace{-0.3cm}
\caption{Event-level mAP comparison on PKU-MMD CS at IoU thresholds 0.1, 0.3, and 0.5. RGB-based methods (top) are listed for reference.}
\vspace{0.1cm}
\label{tab_pku}
\end{table}

\subsection{Ablation Studies}\label{sec:study}
In this section, we conduct a comprehensive ablation study on the contrastive learning strategies, and the impact of the pre-training on PoseCap-1M. 

Firstly, T-MOR relies on a contrastive formulation that coordinates three modalities while preserving the information carried by the pre-extracted visual-textual features. The comparison of loss variants shows that motion and visual-textual cues provide complementary discrimination for action recognition, and text adds useful semantic structure. As shown in Tab.~\ref{tab_ablation}, the best performance is obtained when the contrastive objective jointly promotes alignment among related samples and separation from mismatched ones across two or three modalities.

We also study the number of negative samples and the skeleton backbones. The results (see Appendix) show that a larger queue gives better transfer accuracy, and UNIK provides the best accuracy with lower training time.
\begin{table}[t]
\centering

\begin{center}
\scalebox{0.9}{
\setlength{\tabcolsep}{1.2mm}{
\begin{tabular}{  l c c c c  }

\hline
\multirow{2}*{\textbf{Methods}}&\multirow{2}*{\textbf{Pre-training}} &\multirow{2}*{\textbf{Loss}} &
\multicolumn{1}{c}{\textbf{Smarthome}} & \multicolumn{1}{c}{\textbf{TSU}}\\
& & &\text{CS(\%)} & {\text{ CS(\%)}} \\
\hline
\hline

\text{Baseline}& None & - &\text{24.6}& \text{8.1}  \\
\text{MM}&PoseCap-1M & $\mathcal{L}_{mm}$ &\text{42.5}& \text{12.8}  \\
\text{MV}&PoseCap-1M & $\mathcal{L}_{mv}$ &\text{46.3}& \text{16.3}  \\
\text{MT}&PoseCap-1M & $\mathcal{L}_{mt}$ &\text{49.0}& \text{18.6}  \\

\hline

\text{MM+MV}&Posetics & $\mathcal{L}_{mm}$+$\mathcal{L}_{mv}$ &\text{42.6}& \text{12.5}  \\
\text{MM+MV+MT}&Posetics & $\mathcal{L}_{mm}$+$\mathcal{L}_{mv}$+$\mathcal{L}_{mt}$ &\text{49.6}& \text{18.8}  \\
\hline

\text{MM+MV}&PoseCap-1M & $\mathcal{L}_{mm}$+$\mathcal{L}_{mv}$ &\text{49.3}& \text{19.8}   \\
\text{MM+MT}&PoseCap-1M & $\mathcal{L}_{mm}$+$\mathcal{L}_{mt}$ &\text{51.5}& \text{21.7}   \\
\text{MM+MV+MT}&PoseCap-1M & $\mathcal{L}_{mm}$+$\mathcal{L}_{mv}$+$\mathcal{L}_{mt}$ &\textbf{52.6}& \textbf{23.2}   \\

\hline
\end{tabular}}}

\end{center}
\vspace{-0.3cm}
\caption{Ablation study on Smarthome action classification and TSU action detection in the linear probing setting. V/M/T: Visual/Motion/Text.}
\vspace{-0.15cm}
\label{tab_ablation}
\end{table}

\section{Conclusion}
We presented T-MOR, a motion-aware skeleton representation framework that uses video and language supervision during pre-training to improve human action recognition. Together with PoseCap-1M, the proposed approach improves recognition of complex actions and shows strong few-shot and zero-shot transfer while requiring only skeleton inputs at inference time.
Future work will extend T-MOR pre-training with additional modalities, such as optical flow, to further strengthen action representation learning. 

\vspace{.15cm}
\section{Acknowledgments}
This work was supported in part by the Natural Science Foundation of Jiangsu Province Basic Research Program under Grant BK20250489; in part by the the NSF of China under Grant 62502492; and in part by the French government, through the 
National Research Agency (ANR) with the reference number ANR-23-IACL-0001.

\bibliographystyle{IEEEbib}
\bibliography{icme2026references}

@inproceedings{TGM,
      title={Temporal Gaussian Mixture Layer for Videos},
      booktitle={ICML},
      author={AJ Piergiovanni and Michael S. Ryoo},
      year={2019}
}

@article{graves2005framewise,
  title={Framewise phoneme classification with bidirectional LSTM and other neural network architectures},
  author={Graves, Alex and Schmidhuber, J{\"u}rgen},
  journal={IJCNN},
  year={2005},
  publisher={Elsevier}
}

@article{Shahroudy2016NTURA,
  title={NTU RGB+D: A Large Scale Dataset for {3D} Human Activity Analysis},
  author={Amir Shahroudy and Jun Liu and Tian-Tsong Ng and Gang Wang},
  journal={CVPR},
  year={2016}
}

@article{Yan2018SpatialTG,
  title={Spatial Temporal Graph Convolutional Networks for Skeleton-Based Action Recognition},
  author={S. Yan and Yuanjun Xiong and D. Lin},
  journal={AAAI},
  year={2018},

}

@InProceedings{Carreira_2017_CVPR,
author = {Carreira, Joao and Zisserman, Andrew},
title = {Quo Vadis, Action Recognition? A New Model and the Kinetics Dataset},
booktitle = {CVPR},
year = {2017}
}

@ARTICLE{NTU-120,
  author={J. {Liu} and A. {Shahroudy} and M. {Perez} and G. {Wang} and L. -Y. {Duan} and A. C. {Kot}},
  journal={IEEE TPAMI}, 
  title={NTU RGB+D 120: A Large-Scale Benchmark for {3D} Human Activity Understanding}, 
  year={2020}}

@INPROCEEDINGS{penn,
  author={W. {Zhang} and M. {Zhu} and K. G. {Derpanis}},
  booktitle={ICCV}, 
  title={From Actemes to Action: A Strongly-Supervised Representation for Detailed Action Understanding}, 
  year={2013}}

@INPROCEEDINGS{non-para,
  author={Z. {Wu} and Y. {Xiong} and S. X. {Yu} and D. {Lin}},
  booktitle={CVPR}, 
  title={Unsupervised Feature Learning via Non-parametric Instance Discrimination}, 
  year={2018}
}

@InProceedings{He_2020_CVPR,
author = {He, Kaiming and Fan, Haoqi and Wu, Yuxin and Xie, Saining and Girshick, Ross},
title = {Momentum Contrast for Unsupervised Visual Representation Learning},
booktitle = {CVPR},
year = {2020}
}

@InProceedings{ConNTU,
    author    = {Li, Linguo and Wang, Minsi and Ni, Bingbing and Wang, Hang and Yang, Jiancheng and Zhang, Wenjun},
    title     = {3D Human Action Representation Learning via Cross-View Consistency Pursuit},
    booktitle = {CVPR},
    year      = {2021}
}

@InProceedings{unik,
      title={UNIK: A Unified Framework for Real-world Skeleton-based Action Recognition}, 
      author={Di Yang and Yaohui Wang and Antitza Dantcheva and Lorenzo Garattoni and Gianpiero Francesca and Francois Bremond},
      year={2021},
      booktitle ={BMVC}
}

@inproceedings{guo2022aimclr,
  Title= {Contrastive Learning from Extremely Augmented Skeleton Sequences for Self-supervised Action Recognition},
  Author= {Tianyu, Guo and Hong, Liu and Zhan, Chen and Mengyuan, Liu and Tao, Wang  and Runwei, Ding},
  Booktitle= {AAAI},
  Year= {2022}
}

@article{yang2022via,
      title={ViA: View-invariant Skeleton Action Representation Learning via Motion Retargeting}, 
      author={Di Yang and Yaohui Wang and Antitza Dantcheva and Lorenzo Garattoni and Gianpiero Francesca and Francois Bremond},
      year={2024},
      journal={IJCV}
}

@INPROCEEDINGS{motion,
  author={Diba, Ali and Sharma, Vivek and Van Gool, Luc and Stiefelhagen, Rainer},
  booktitle={ICCV}, 
  title={DynamoNet: Dynamic Action and Motion Network},
  year={2019},
 }

@ARTICLE{Dai_2022_PAMI,
  author={Dai, Rui and Das, Srijan and Sharma, Saurav and Minciullo, Luca and Garattoni, Lorenzo and Bremond, Francois and Francesca, Gianpiero},
  journal={IEEE TPAMI}, 
  title={Toyota Smarthome Untrimmed: Real-World Untrimmed Videos for Activity Detection}, 
  year={2022},
  }

@article{liu2017pku, 
  title={PKU-MMD: A Large Scale Benchmark for Continuous Multi-Modal Human Action Understanding},
  author={Chunhui, Liu and Yueyu, Hu and Yanghao, Li and Sijie, Song and Jiaying, Liu},
  journal={arXiv:1703.07475},
  year={2017}
}

@inproceedings{jcrnn,
  author = {Li, Yanghao and Lan, Cuiling and Xing, Junliang and Zeng, Wenjun and Yuan, Chunfeng and Liu, Jiaying},
  title = {Online Human Action Detection using Joint Classification-Regression Recurrent Neural Networks},
  booktitle = {ECCV},
  year = {2016},
}

@inproceedings{skeletonbox,
  author = {Li, Bo and Chen, Huahui and Chen, Yucheng and Dai, Yuchao and He, Mingyi},
  title = {Skeleton Boxes: Solving skeleton based action detection with a single deep convolutional neural network},
  booktitle = {ICMEW},
  year = {2017},
}

@inproceedings{windowproposal,
	year = 2017,
	booktitle = {ICMEW},
    author = {Chuankun Li and Yonghong Hou and Pichao Wang and Wanqing Li},
	title = {Joint Distance Maps Based Action Recognition With Convolutional Neural Networks},
}

@inproceedings{GRU-GD,
  author = {Luo, Zelun and Hsieh, Jun-Ting and Jiang, Lu and Niebles, Juan Carlos and Fei-Fei, Li},
  title = {Graph Distillation for Action Detection with Privileged Modalities},
  booktitle = {ECCV},
  year = {2018},
}

@InProceedings{Dai_2021_ICCV,
    author    = {Dai, Rui and Das, Srijan and Bremond, Fran\c{c}ois},
    title     = {Learning an Augmented RGB Representation With Cross-Modal Knowledge Distillation for Action Detection},
    booktitle = {ICCV},
    year      = {2021},
}

@inproceedings{tevet2022motionclip,
title={MotionCLIP: Exposing Human Motion Generation to CLIP Space},
author={Tevet, Guy and Gordon, Brian and Hertz, Amir and Bermano, Amit H and Cohen-Or, Daniel},
booktitle={ECCV},
year={2022}
}

@inproceedings{radford2021clip,
      title={Learning Transferable Visual Models From Natural Language Supervision}, 
      author={Alec Radford and Jong Wook Kim and Chris Hallacy and Aditya Ramesh and Gabriel Goh and Sandhini Agarwal and Girish Sastry and Amanda Askell and Pamela Mishkin and Jack Clark and Gretchen Krueger and Ilya Sutskever},
      year={2021},
      booktitle={ICML}
}

@article{xu2021vlm-videoclip,
      title={VLM: Task-agnostic Video-Language Model Pre-training for Video Understanding}, 
      author={Hu Xu and Gargi Ghosh and Po-Yao Huang and Prahal Arora and Masoumeh Aminzadeh and Christoph Feichtenhofer and Florian Metze and Luke Zettlemoyer},
      year={2021},
      journal={arXiv preprint arXiv:2105.09996},
}

@InProceedings{Mao_2023_ICCV,
    author    = {Mao, Yunyao and Deng, Jiajun and Zhou, Wengang and Fang, Yao and Ouyang, Wanli and Li, Houqiang},
    title     = {Masked Motion Predictors are Strong 3D Action Representation Learners},
    booktitle = {ICCV},
    year      = {2023}
}

@article{wang2022internvideo,
  title={InternVideo: General Video Foundation Models via Generative and Discriminative Learning},
  author={Wang, Yi and Li, Kunchang and Li, Yizhuo and He, Yinan and Huang, Bingkun and Zhao, Zhiyu and Zhang, Hongjie and Xu, Jilan and Liu, Yi and Wang, Zun and Xing, Sen and Chen, Guo and Pan, Junting and Yu, Jiashuo and Wang, Yali and Wang, Limin and Qiao, Yu},
  journal={arXiv:2212.03191},
  year={2022}
}

@INPROCEEDINGS{vit,
    title   = {An Image is Worth 16x16 Words: Transformers for Image Recognition at Scale},
    author  = {Alexey Dosovitskiy and Lucas Beyer and Alexander Kolesnikov and Dirk Weissenborn and Xiaohua Zhai and Thomas Unterthiner and Mostafa Dehghani and Matthias Minderer and Georg Heigold and Sylvain Gelly and Jakob Uszkoreit and Neil Houlsby},
    year    = {2021},
    booktitle = {ICLR},
}

@INPROCEEDINGS{wang2023internvid,
  title={InternVid: A Large-scale Video-Text Dataset for Multimodal Understanding and Generation},
  author={Wang, Yi and He, Yinan and Li, Yizhuo and Li, Kunchang and Yu, Jiashuo and Ma, Xin and Chen, Xinyuan and Wang, Yaohui and Luo, Ping and Liu, Ziwei and Wang, Yali and Wang, Limin and Qiao, Yu},
  booktitle={ICLR},
  year={2024}
}

@INPROCEEDINGS{wang2023videomaev2,
    title={VideoMAE V2: Scaling Video Masked Autoencoders With Dual Masking},
    author={Wang, Limin and Huang, Bingkun and Zhao, Zhiyu and Tong, Zhan and He, Yinan and Wang, Yi and Wang, Yali and Qiao, Yu},
    booktitle={CVPR},
    year= {2023}
}

@inproceedings{Ma2022XCLIP,
    title={{X-CLIP:}: End-to-End Multi-grained Contrastive Learning for Video-Text Retrieval},
    author={Yiwei Ma and Guohai Xu and Xiaoshuai Sun and Ming Yan and Ji Zhang and Rongrong Ji},
    booktitle={ACMMM},
    year={2022}
}

@inproceedings{Maaz2023VideoChatGPT,
    title={Video-ChatGPT: Towards Detailed Video Understanding via Large Vision and Language Models},
    author={Maaz, Muhammad and Rasheed, Hanoona and Khan, Salman and Khan, Fahad Shahbaz},
    booktitle={ACL},
    year={2024}
}

@inproceedings{chakraborty2024llavidal,
      title={LLAVIDAL: Benchmarking Large Language Vision Models for Daily Activities of Living}, 
      author={Rajatsubhra Chakraborty and Arkaprava Sinha and Dominick Reilly and Manish Kumar Govind and Pu Wang and Francois Bremond and Srijan Das},
      booktitle={CVPR},
      year={2025},
}

@article{li2024llavaonevisioneasyvisualtask,
      title={LLaVA-OneVision: Easy Visual Task Transfer}, 
      author={Bo Li and Yuanhan Zhang and Dong Guo and Renrui Zhang and Feng Li and Hao Zhang and Kaichen Zhang and Yanwei Li and Ziwei Liu and Chunyuan Li},
      journal={arXiv preprint arXiv:2408.03326},
      year={2024},
}

@article{lin2023videollavalearningunitedvisual,
      title={Video-LLaVA: Learning United Visual Representation by Alignment Before Projection}, 
      author={Bin Lin and Yang Ye and Bin Zhu and Jiaxi Cui and Munan Ning and Peng Jin and Li Yuan},
      journal={arXiv preprint arXiv:2311.10122},
      year={2023},
}

@InProceedings{ali2024eccv,
      title={Are Visual-Language Models Effective in Action Recognition? A Comparative Study}, 
      author={Mahmoud Ali and Di Yang and François Brémond},
      year={2024},
      booktitle={ECCVW}, 
}

@inproceedings{motionbert2022,
  title={MotionBERT: A Unified Perspective on Learning Human Motion Representations}, 
  author={Zhu, Wentao and Ma, Xiaoxuan and Liu, Zhaoyang and Liu, Libin and Wu, Wayne and Wang, Yizhou},
  booktitle={ICCV},
  year={2023},
}

@ARTICLE{language,
  author={He, Tian and Chen, Yang and Gao, Xu and Wang, Ling and Hu, Ting and Cheng, Hong},
  journal={IEEE TCSVT}, 
  title={Enhancing Skeleton-Based Action Recognition With Language Descriptions From Pre-Trained Large Multimodal Models}, 
  year={2025},
}

@inproceedings{ali2024,
  TITLE = {{Quo Vadis, Video Understanding with Vision-Language Foundation Models?}},
  AUTHOR = {Ali, Mahmoud and Yang, Di and Sinha, Arkaprava and Reilly, Dominick and Das, Srijan and Francesca, Gianpiero and Bremond, Francois},
  BOOKTITLE = {{NeurIPSW}},
  YEAR = {2024},
}

@inproceedings{ester1996dbscan,
  title={A Density-Based Algorithm for Discovering Clusters in Large Spatial Databases with Noise},
  author={Ester, Martin and Kriegel, Hans-Peter and Sander, J{\"o}rg and Xu, Xiaowei},
  booktitle={KDD},
  year={1996}
}

@article{li2026stnmamba,
  title={Stnmamba: Mamba-based spatial-temporal normality learning for video anomaly detection},
  author={Li, Zhangxun and Zhao, Mengyang and Yang, Xuan and Liu, Yang and Sheng, Jiamu and Zeng, Xinhua and Wang, Tian and Wu, Kewei and Jiang, Yu-Gang},
  journal={IEEE TMM},
  year={2026},
}

@article{liu2025crcl,
  title={CRCL: Causal Representation Consistency Learning for Anomaly Detection in Surveillance Videos},
  author={Liu, Yang and Wang, Hongjin and Wang, Zepu and Zhu, Xiaoguang and Liu, Jing and Sun, Peng and Tang, Rui and Du, Jianwei and Leung, Victor and Song, Liang},
  journal={IEEE TIP},
  year={2025}
}

@article{liu2025anomaly,
  title={Anomaly detection and generation with diffusion models: A survey},
  author={Liu, Yang and Liu, Jing and Li, Chengfang and Xi, Rui and Li, Wenchao and Cao, Liang and Wang, Jin and Yang, Laurence T and Yuan, Junsong and Zhou, Wei},
  journal={arXiv:2506.09368},
  year={2025}
}

@article{liu2025privacy,
  title={Privacy-preserving video anomaly detection: A survey},
  author={Liu, Yang and Liu, Siao and Zhu, Xiaoguang and Li, Jielin and Yang, Hao and Teng, Liangyu and Guo, Juncen and Wang, Yan and Yang, Dingkang and Liu, Jing},
  journal={IEEE TNNLS},
  year={2025},
}

@phdthesis{yang2024learning,
  title={Learning effective video representations for action recognition},
  author={Yang, Di},
  year={2024},
  school={Universit{\'e} C{\^o}te d'Azur}
}

\end{document}